\pdfoutput=1

\documentclass[11pt]{article}

\usepackage{acl}

\usepackage{times}
\usepackage{latexsym}

\usepackage[T1]{fontenc}

\usepackage[utf8]{inputenc}

\usepackage{microtype}

\usepackage{scalerel}
\usepackage{booktabs}
\usepackage{longtable}
\usepackage{amsmath}
\usepackage{amssymb}

\definecolor{green(munsell)}{rgb}{0.0, 0.66, 0.47}

\title{A Multi-Document Coverage Reward\\ for RELAXed Multi-Document Summarization}

\author{Jacob Parnell\textsuperscript{1,2}, Inigo Jauregi Unanue\textsuperscript{1,2}, Massimo Piccardi\textsuperscript{1} \\
  \textsuperscript{1}University of Technology Sydney, NSW, Australia \\
  \textsuperscript{2}RoZetta Technology, NSW, Australia \\
  \{\texttt{jacob.parnell,inigo.jauregi}\}\texttt{@rozettatechnology.com} \\
  \texttt{massimo.piccardi@uts.edu.au} \\}

\begin{document}
\maketitle
\begin{abstract}
Multi-document summarization (MDS) has made significant progress in recent years, in part facilitated by the availability of new, dedicated datasets and capacious language models. However, a standing limitation of these models is that they are trained against limited references and with plain maximum-likelihood objectives. As for many other generative tasks, reinforcement learning (RL) offers the potential to improve the training of MDS models; yet, it requires a carefully-designed reward that can ensure appropriate leverage of both the reference summaries and the input documents. For this reason, in this paper we propose fine-tuning an MDS baseline with a reward that balances a reference-based metric such as ROUGE with coverage of the input documents. To implement the approach, we utilize \textit{RELAX} \citep{grathwohl2018backpropagation}, a contemporary gradient estimator which is both low-variance and unbiased, and we fine-tune the baseline in a few-shot style for both stability and computational efficiency. Experimental results over the Multi-News and WCEP MDS datasets show significant improvements of up to $+0.95$ pp average ROUGE score and $+3.17$ pp METEOR score over the baseline, and competitive results with the literature. In addition, they show that the coverage of the input documents is increased, and evenly across all documents.
\end{abstract}

\section{Introduction}
\label{sec:intro}

Multi-document summarization (MDS) aims to consolidate salient points of information across a set of documents into a concise summary. The main requirement for the summary is that it adequately represent the document set, with low redundancy and high coverage across all documents, while at the same time being readable and fluent. Combined with this, is the need to develop techniques that can handle the significant memory complexity required to tackle MDS. Recently, the release of dedicated datasets \cite{fabbri2019, gholipour-ghalandari-etal-2020-large}, and intelligently designed  Transformer models \cite{j.2018generating,liu-lapata-2019-hierarchical, Beltagy2020Longformer}, have helped drive advancements in multi-document summarization, generally improving the accuracy and fluency of the predicted summaries.
However, aspects such as the requirement to cover as much salient information from the input documents as possible, whilst still maintaining low repetition and low redundancy, have certainly been less explored to date \citep{nayeem-etal-2018-abstractive, mao-etal-2020-multi}. 

Within the sphere of contemporary neural MDS models, two main lines of investigation can be identified: graph-based approaches \cite{li-etal-2020-leveraging-graph, pasunuru-etal-2021-efficiently}, and concatenation approaches \cite{j.2018generating, zhang2020}. The former are approaches that rely on the construction of graphs to capture the inter- and intra-document relations. While powerful, they need to elicit the relations explicitly. The latter instead assume that all the input documents within a document set can be simply concatenated, possibly with document separators and tags, such that the relations can be ``discovered'' by the model. 
Like ordinary summarization, also MDS comes in two remarkably different styles: \textit{extractive}, where the generated summaries consist of verbatim sentences from the original input documents \citep{nallapati2017}, and \textit{abstractive}, where the model is instead encouraged to generate a paraphrased understanding of the input documents. The intrinsic appeal of abstractive summaries and the advent of sequence-to-sequence models have increasingly shifted the trend toward abstractive summarization \citep{see-etal-2017-get, paulus2017deep, fabbri2019, lewis2020, zhang2020}. As for what models are concerned, abstractive MDS has made increasing use of transformers, both ``conventional'' \cite{lewis2020, zhang2020} and modified to accommodate the characteristic input length of multi-document sets \citep{Beltagy2020Longformer, zaheerbigbird2020}.  


Similarly to general summarization, the majority of MDS models are trained using the negative log-likelihood (NLL) as training objective, which aims to maximize the conditional log-likelihood of the tokens of a given reference summary. Despite its speed and efficacy, the NLL exhibits both the \textit{wrong-objective} problem \cite{ding2017coldrl}, where the model is trained on a convenient objective rather than a desirable one, and the well-known \textit{exposure bias} problem \cite{bengio2015, ranzato2016sequence}. 
To alleviate these issues, reinforcement learning has been adopted in summarization, as in other language generation tasks, to train the model with a more appropriate objective  \cite{li2019dsr, parnell-etal-2021-rewardsofsum}. However, its effective use for MDS requires a reward function that can appropriately balance the reference summary and the multiple input documents in the document set. For this reason, in this paper we propose exploring a reward that combines a reference-based metric such as ROUGE with a coverage term over the input documents.
To implement the reinforcement learning approach, we employ a contemporary gradient estimator of the policy gradient, RELAX \citep{grathwohl2018backpropagation}, which is both low-variance and unbiased. In addition, to limit the computation and the risk of parameter drift, we apply the objective to fine-tune an NLL-pretrained model in a few-shot manner.
In light of the above, this paper makes the following contributions:

\begin{enumerate}

    \item a reward for reinforcement learning that combines a ROUGE score and a multi-document coverage score, to simultaneously adhere to both the reference summaries and the input documents;
    
    \item a reinforcement learning implementation that leverages a low-variance and unbiased gradient estimator of the policy gradient, RELAX;
    
    \item experimental results and a comprehensive analysis over two MDS datasets (Multi-News and WCEP), showing the empirical effectiveness of the proposed approach. 
    
\end{enumerate}

The rest of this paper is organized as follows: first the related work is reviewed in Section \ref{sec:related}, and then the proposed approach is introduced in Section \ref{sec:proposed_approach}. Section \ref{sec:experiments} describes the experimental set-up and main results, while Section  \ref{sec:analysis} presents a more detailed analysis of the main components of the proposed approach. Eventually, Section \ref{sec:conclusion} summarizes our findings and concludes the paper.


\section{Related Work}
\label{sec:related}

Early work in multi-document summarization (MDS) that pre-dates the neural era \citep{mani-bloedorn-97, erkan-and-radev04, christensen-etal-2013-towards} was shaped around the notion of MDS as a collection of graph structures. As approaches in language generation naturally evolved into neural-based \citep{rush2015, ranzato2016sequence}, later improved with the emergence of large, pre-trained language models \citep{devlin2018, lewis2020, zhang2020},
the effort shifted to integrating these graph structures into the models, often building on top of strong single-document summarization (SDS) baselines  \citep{lebanoff2018, zhang-etal-2018-adapting}.


Concurrently, the growing  interest in multi-document summarization has led to the development of dedicated, multi-document datasets such as WikiSum \citep{j.2018generating},  Multi-News \citep{fabbri2019}, Wikipedia Current Events Portal (WCEP) \cite{gholipour-ghalandari-etal-2020-large} and others. The typical amount of input data that comes with these datasets has increased the pressure on the models to be able to handle larger inputs. For instance, WCEP has up to 100 documents in each document set, and 63.7 on average. As such, the standard transformers used to develop successful SDS models such as BART \citep{lewis2020} and PEGASUS \citep{zhang2020} have proved inadequate for MDS due to their limited maximum input length (in the order of $10^{3}$ tokens) and quadratic memory complexity \citep{Beltagy2020Longformer}. In turn, this has prompted the development of long transformer models such as Longformer \citep{Beltagy2020Longformer} (built upon BART) and BigBird \citep{zaheerbigbird2020} (built upon PEGASUS) which, thanks to their smart attention layers that scale linearly with the input length, have opened up the possibility of presenting the input documents ``at once'', allowing these re-designed attention mechanisms to discover both inter- and intra-document relations. 

Document summarization, as have other language generation tasks, has often been criticized for using maximum-likelihood training objectives that may prove limitative for the eventual performance of the models \citep{ding2017coldrl}. For this reason, reinforcement learning has been employed as an alternative, to directly optimize the models over evaluation metrics and explicitly reward the quality of the model's predictions. 
Reinforcement learning approaches have used metrics such as ROUGE-1, ROUGE-2 and ROUGE-L F1 \citep{paulus2017deep}, and also more contemporary scoring functions such as BERTScore \citep{zhang2020bertscore} as rewards, often mixed with maximum-likelihood objectives. When applying reinforcement learning to MDS, we contend that the reward should not simply be a ROUGE score against the reference summary, since this would dismiss key characteristics of the task such as inter-document information transfer. For instance,  \citet{mao-etal-2020-multi} have leveraged maximal marginal relevance \citep{carbonell1998} to mollify higher-order information redundancy between the input documents. Several other performance measures could potentially be included in the reward, such as extractive fragment coverage and density \citep{grusky2018newsroom} and MINT \citep{dreyer2021}, but to the best of our knowledge they have never been utilized as, or for, training objectives.

To address this gap, in this paper we propose leveraging a modified coverage reward to improve information coverage across all the documents in the input set, jointly with a principled policy gradient estimator (RELAX) and a performing long transformer model (the BART Longformer Encoder-Decoder, or BART-LED), in the hope of benefiting from the synergy between these components.


\section{Proposed Approach}
\label{sec:proposed_approach}


In this section, we present the details of the proposed approach, including the reinforcement learning framework (Section \ref{subsec:rl_grad_est}), the multi-document coverage reward (Section \ref{subsec:coverage}), and the overall training objective (Section \ref{subsec:final_train_obj}).

\subsection{Reinforcement Learning Gradient Estimators}
\label{subsec:rl_grad_est}

Given a set of documents in input, simply noted as $x$, and a summary with $T$ tokens, $y = \{y_1, \dots, y_T\}$, the predictive distribution, also known as policy in reinforcement learning, can be noted as $p(y_{t}|y_{1},\ldots,y_{t-1}, x)$. The policy gradient theorem \cite{sutton1999policy} states that an estimator for the gradient of the reinforcement learning risk can be expressed as:

\vspace{-12pt}

\begin{equation}
\label{eq:grad}
\begin{aligned}
\Delta = -r \sum\limits_{t=1}^{T} \: \frac{\partial}{\partial \theta} \log p(y_{t}^{s}|y_{1}^{s},\ldots,y_{t-1}^{s}, x) \\
\end{aligned}
\end{equation}

\noindent where $y^s_1, \ldots, y^s_T$ is a sequence sampled from the policy, $r$ is a function that rewards its quality, and $\theta$ collectively denotes all the policy's parameters. This estimator is the well-known REINFORCE \cite{williams92reinforce} and is a baseline of reinforcement learning. At its turn, the gradient can be easily turned into a loss function to be used with automatic differentiation:

\begin{equation}
\label{eq:reinforce}
\begin{aligned}
L_{\scaleto{REINFORCE}{3pt}} &= -r \sum\limits_{t=1}^{T} \: \log p(y_{t}^{s}|y_{1}^{s},\ldots,y_{t-1}^{s}, x) \\
&= -r \log p(y^s)\\
\end{aligned}
\end{equation}

The sampled sequence in (\ref{eq:reinforce}), $y^s = \{y_1^s, \dots, y_T^s\}$, can be obtained with any usual sampling approach such as teacher-forcing, student-forcing, or scheduled sampling \citep{bengio2015}. While the samples can be drawn from a standard categorical distribution, in our experiments we utilize the Gumbel-Softmax  re-parameterization \cite{jang2017categorical} to obtain the categorical samples from transformed samples of a Gumbel distribution. The reason for the re-parameterization is that the Gumbel-Softmax samples are needed for the RELAX estimator that we introduce in the following. For a generic sample, $y_t^s$, the re-parameterization can be concisely expressed as:

\begin{equation}
\label{eq:Gumbel-Softmax}
\begin{split}
&y_t^s = \text{argmax}(z_t)\\
&z_t \sim \text{Gumbel-Softmax}(p_t,\tau)
\end{split}
\end{equation}

\noindent where $z_t$ is a Gumbel-Softmax sample of size equal to that of the vocabulary that acts as a ``soft'' prediction, \ $p_t$ is the probability vector over the vocabulary at slot $t$, $\tau$\ is a temperature parameter controlling the sparsity of $z_t$, and $\text{argmax}(z_t)$ returns the index of $z_t$'s largest value. This re-parameterization is provenly equivalent to directly sampling $y_t^s$ from $\text{Cat}(p_t)$ (the reader can refer to \citet{jang2017categorical} for details). 




REINFORCE is an unbiased estimator of the theoretical gradient, but it typically suffers from a high variance which can affect the convergence and effectiveness of training. To curb its high variance, techniques based on control variates and the subtraction of simple baselines have been proposed and even applied to summarization \citep{rennie2017selfcritical, paulus2017deep}. However, our early experiments showed that these approaches were not promising for the given task. In addition, some of these estimators introduce a ``bias'', i.e. a mean difference with respect to the theoretical gradient. More recently, the RELAX gradient estimator has been shown to empirically outperform REINFORCE, thanks to its ability to reduce the variance while remaining unbiased  \citep{grathwohl2018backpropagation}. The corresponding RELAX loss can be expressed as: 

\begin{equation}
\label{eq:relax}
\begin{aligned}
L_{\scaleto{RELAX}{3pt}} = -[r - c_{\phi}(\tilde{z})] \log p(y^s) + c_{\phi}(z) - c_{\phi}(\tilde{z})\\
\end{aligned}
\end{equation}

In (\ref{eq:relax}), $c_{\phi}(\tilde{z})$ is a control variate of parameters $\phi$ which is expected to correlate tightly with the reward to reduce the variance, and term $c_{\phi}(z) - c_{\phi}(\tilde{z})$ ensures that the overall gradient remains an unbiased estimator of the theoretical gradient. Variable $z = \{z_1, \ldots, z_T \}$ denotes the sequence of the Gumbel-Softmax samples, while variable $\tilde{z}$ denotes the sequence of samples from a Gumbel-Softmax distribution conditioned on the observed values of $y^s$.
Operationally, $z_t$ is sampled first, unconditionally, then $y_t^s$ is derived with the argmax, and finally $\tilde{z_t}$ is sampled from a suitably conditioned Gumbel-Softmax distribution; details can be found in \citet{grathwohl2018backpropagation}, Appendix B - Categorical. Overall, the RELAX estimator is both unbiased and low-variance. 




The control variate in our experiments is a simple two-layer feed-forward network that is constructed to correlate with the ROUGE scoring function. We obtain this by feeding the concatenation of the soft predictions, $z$ (or, in turn, $\tilde{z}$), and the reference summary, $y$, as input to the control variate. This allows the model to learn to score the soft predictions and their targets in a way that mimics the ROUGE prediction-reference score. In detail, the architecture consists of two fully-connected linear layers, each followed by a ReLU linear activation function, and a final sigmoid activation function that normalizes the output of the last layer. Eventually, the output of the sigmoid is averaged to produce the control variate.\footnote{We release our code to permit complete reproducibility of our experiments: \url{https://github.com/jacob-parnell-rozetta/longformer_coverage/}} 


\subsection{Multi-Document Coverage Reward}
\label{subsec:coverage}


The design of an effective reward is another key aspect of a reinforcement learning objective. In our work, we have aimed to design an overall reward that could simultaneously remain faithful to: a) the reference summary, to ensure adequate generation performance, and b) the input documents, to cover as many important details as possible, and hopefully, support generalization. Relying solely on the reference summaries, given the large input size, does not seem to promise sufficient guidance, and our experiments have confirmed that. To implement the reward, we have chosen to use ROUGE-L F1 for the references and a multi-document coverage score for the input documents that we describe hereafter.

Several quantitative measures of coverage exist in the literature, and have found ample use in describing the properties of summarization datasets and the performance of models. For our work, we have adopted the \textit{extractive fragment coverage (EFC)} of  \citet{grusky2018newsroom}. The EFC measures the percentage of words in a summary that are part of ``extractive fragments'' within an input document, which are simply multi-word phrases shared between the input document and the summary. It is a simple precision-type measurement that looks at how much of the prediction is in the input document. Noting an individual document as $D$, a summary as $y$ and an extractive fragment as $f$, the EFC can be expressed as:  

\begin{equation}
\label{eq:coverage}
\begin{aligned}
EFC(y,D) = \frac{1}{|y|}\sum_{f \in \mathcal{F}(y,D)} |f|
\end{aligned}
\end{equation}

\noindent where the $|\cdot|$ operator is used to denote length. To promote an even improvement in coverage across the input documents, we propose a multi-document extension of the EFC that reaches its highest value when the coverage across the input documents is evenly distributed. Let us note the input document set here as $\mathcal{D}$, and the EFC coverage vector over the document set as $cov(y,\mathcal{D})$. We also note the sample mean of a vector $x$ as $\mu(x)$, the sample standard deviation as $\sigma(x)$, and their ratio (the inverse coefficient of variation) as $c_v^{-1}(x)$. This allows us to compute a ``normalized'' coverage score for a summary,  $c_v^{-1}(cov(y,\mathcal{D}))$, which takes larger values the more the scores are uniform across the document set. In addition, inspired by \citet{kryscinski2018abstraction}, we define a reward that pits the normalized coverage score of the prediction, $y^s$, against that of the reference, $y$:

\begin{equation}
\label{eq:coverage_reward}
\begin{aligned}
r^{cov} = \frac{c_v^{-1}(cov(y^s,\mathcal{D})) - c_v^{-1}(cov(y,\mathcal{D}))}{c_v^{-1}(cov(y^s,\mathcal{D}))}
\end{aligned}
\end{equation}

Eventually, to ensure that short summaries are not unfairly rewarded with high coverage scores, we normalize the reward by the length ratio of the prediction and the reference:

\begin{equation}
\label{eq:final coverage_reward}
\begin{aligned}
\hat{r}^{cov} = r^{cov} \frac{|y^s|}{|y|}
\end{aligned}
\end{equation}

Overall, the $\hat{r}^{cov}$ reward regards a prediction as ``good'' if it enjoys high average coverage of the input documents, the coverage is evenly distributed, and the prediction is of sufficient length. The reference summary acts as a baseline, making the reward additive if the prediction outperforms the reference, and subtractive if otherwise.


Since ROUGE-L F1 and the coverage reward are not necessarily up to scale, to obtain the final reward, $r$, we perform a convex combination with a scaling coefficient, $\beta$:

\begin{equation}
\label{eq:cov_reward+rouge}
\begin{aligned}
r =  \text{ROUGE-L F1}(y^s,y) + \beta \hspace{2pt} \hat{r}^{cov} \\ \end{aligned}
\end{equation}

\subsection{Overall Training Objective}
\label{subsec:final_train_obj}
As training strategy, we first train the model with the negative log-likelihood and choose the best model with a criterion based on the validation performance. After that, the model is fine-tuned with the reinforcement learning objective. In many past works, the reinforcement learning objective has been used mixed with the NLL for stability \citep{paulus2017deep, li2019dsr, parnell-etal-2021-rewardsofsum}. However, we assume that the model has already ``warmed up'' to the training data during its NLL pre-training stage, and only use either $L_{REINFORCE}$  (\ref{eq:reinforce}) or $L_{RELAX}$  (\ref{eq:relax}) for fine-tuning. To prevent excessive drifting from the NLL pre-trained model, we limit the fine-tuning to a few ($\approx 1,000$) shots and a relatively low learning rate ($3 \times 10^{-6}$).


\section{Experiments}
\label{sec:experiments}

\subsection{Datasets}
We have carried out multiple experiments over two MDS datasets in the news domain: Multi-News \cite{fabbri2019} and Wikipedia Current Events Portal (WCEP) \cite{gholipour-ghalandari-etal-2020-large}. For WCEP, we specifically use the WCEP-100 version, which exclusively limits the number of articles within a document set to 100. We have chosen these datasets as they cover an ample spread of summary lengths and numbers of input documents, with Multi-News having longer reference summaries on average.  Appendix \ref{subsec:appendix-dataset-stats} reports the datasets' main statistics as presented in the original papers \citep{fabbri2019, gholipour-ghalandari-etal-2020-large}\footnote{Instructions to access the datasets are available in Appendix \ref{subsec:appendix-dataset-stats}}.

\subsection{Evaluation Metrics}
Like most previous works, we use the F1 variants of the ROUGE-\textit{N} scores\footnote{\url{https://pypi.org/project/rouge-score/}} \cite{lin2004} for performance evaluation. In our use of ROUGE, we choose not to stem the predictions and the references during scoring. Since we use the ROUGE-L F1 score in our reward, to avoid circularity we also include METEOR\footnote{\url{http://www.nltk.org/_modules/nltk/translate/meteor_score.html}} \cite{lavie-agarwal-2007-meteor} in the performance evaluation. Differently from our ROUGE implementation, METEOR uses stemming, synonyms, and other paraphrastic matching in the \textit{n}-gram matching stage. In a recent study, both ROUGE and METEOR have displayed high correlation with a number of desirable summarization properties such as coherence, consistency, fluency, and relevance \citep{summeval2021}.

\subsection{Main Settings}
We have implemented our approach on top of BART-LED \citep{Beltagy2020Longformer}. We utilize the generous maximum encoding length (16384 tokens) of this long-input transformer, by concatenating all the documents in a document set to form a single input to the model. The individual documents are separated by an \texttt{[END]} token, and the input is truncated to the maximum length. For every experiment, we report the average of three independently-initialized training runs. For each result, we have also run a nonparametric bootstrap test for statistical significance, and highlighted the results that are significantly different from the baseline. In the reward, the $\beta$ hyperparameter has been set to 1.0 with a validation described in Appendix \ref{subsec:appendix-scaling-coverage-term-analysis}. All other hyperparameters are described in Appendix \ref{subsec:appendix-modelhyp}.

\begin{table*}[!ht]
\centering
\small
\begin{tabular}{cccccc}
\hline
\textbf{Model} & \textbf{R-1} & \textbf{R-2} & \textbf{R-L} & \textbf{METEOR} \\
\hline
Previous Work \\
\hline
HiMAP \cite{fabbri2019} & 44.17 & 16.05 & 21.38 & - \\
Hierarchical Transformer \cite{liu-lapata-2019-hierarchical} & 42.36 & 15.27 & 22.08 & -\\
GraphSum \cite{li-etal-2020-leveraging-graph} & 45.02 & 16.69 & 22.50 & - \\
GraphSum + RoBERTa \cite{li-etal-2020-leveraging-graph} & 45.87 & 17.56 & 23.39 & - \\
BART-Long \cite{pasunuru-etal-2021-efficiently} & \textbf{48.54} & 18.56 & 23.78 & - \\
\hline
 Our Models \\
\hline
BART-LED (Baseline) & 46.89 & 18.50 & 24.84 & 29.61 \\
ROUGE-L + REINFORCE & 46.52 & 18.49 & 24.91 & 29.19 \\
ROUGE-L + Coverage ($\beta=1.0$) + REINFORCE & 46.39 & 18.29 & 24.74 & 29.02 \\
ROUGE-L + RELAX & 47.05{$^\dagger$} & 18.76{$^\dagger$} & 24.99{$^\dagger$} & 29.98{$^\dagger$} \\
ROUGE-L + Coverage ($\beta=1.0$) + RELAX & 47.23{$^\dagger$} & \textbf{18.86}{$^\dagger$} & \textbf{25.03}{$^\ddag$} & \textbf{30.53}{$^\dagger$} \\
\hline
\end{tabular}
\caption{\label{mn_results}
Average ROUGE and METEOR scores over the Multi-News test set. ($\dagger$) and ($\ddag$) refer to statistically significant differences with respect to our baseline with a $p$-value < 0.01 and < 0.05, respectively, in a bootstrap hypothesis test \cite{dror2018}. The best scores are bolded.
}
\end{table*}

\begin{table*}[!ht]
\centering
\small
\begin{tabular}{cccccc}
\hline
\textbf{Model} & \textbf{R-1} & \textbf{R-2} & \textbf{R-L} & \textbf{METEOR} \\
\hline
Previous Work \\
\hline
TSR \cite{gholipour-ghalandari-etal-2020-large} & 35.30 & 13.70 & 25.70 & - \\
BERTReg \cite{gholipour-ghalandari-etal-2020-large} & 35.00 & 13.50 & 25.50 & - \\
Submodular+ABS \cite{gholipour-ghalandari-etal-2020-large} & 34.40 & 13.10 & 25.00 & - \\
BART-WCEP-DynE-5 \cite{hokamp2020} & 35.40 & 15.10 & 25.60 & - \\
\hline
 Our Models \\
\hline
BART-LED (Baseline) & 39.79 & 18.94 & 32.10 & 29.04 \\
ROUGE-L + REINFORCE & 40.25{$^\dagger$} & 18.18 & 31.58 & 30.91{$^\dagger$} \\
ROUGE-L + Coverage ($\beta=1.0$) + REINFORCE & 40.68{$^\dagger$} & 18.80 & 32.71{$^\ddag$} & 30.28{$^\ddag$} \\
ROUGE-L + RELAX & \textbf{41.11}{$^\dagger$} & \textbf{19.46}{$^\dagger$} & \textbf{33.13}{$^\dagger$} & 30.57{$^\dagger$} \\
ROUGE-L + Coverage ($\beta=1.0$) + RELAX & 40.78{$^\dagger$} & 19.14 & 32.37 & \textbf{32.21}{$^\dagger$} \\
\hline
\end{tabular}
\caption{\label{wcep_results}
Average ROUGE and METEOR scores over the WCEP test set. ($\dagger$) and ($\ddag$) refer to statistically significant differences with respect to our baseline with a $p$-value < 0.01 and < 0.05, respectively, in a bootstrap hypothesis test \cite{dror2018}. The best scores are bolded.
}
\end{table*}

\subsection{Results}
\label{sec:results}


\textbf{Multi-News}.
Table \ref{mn_results} compares the results over the Multi-News test set for the baseline, our proposed approaches and previous work from the literature. We first note that our BART-LED model has performed as a strong baseline, with its results being comparable to those of BART-Long \citep{pasunuru-etal-2021-efficiently}, which is based on the same BART Longformer architecture. In detail, BART-Long has reported a higher ROUGE-1 score, our baseline has reported a higher ROUGE-L score, and both have reported similar ROUGE-2 scores. Therefore, we regard our performance as comparable on the whole, with the differences most likely due to different hyperparameters. 

Amongst our results, the models fine-tuned with REINFORCE have achieved worse results than the baseline. This is evidence that a vanilla implementation of the policy gradient is not necessarily better than a standard NLL objective. Conversely, the models fine-tuned with RELAX have surpassed both the NLL baseline and virtually all the previous work. The best results have been achieved with the inclusion of the coverage term, with an improvement of $+0.36$ ROUGE-2 pp over the NLL baseline and a marked improvement of $+0.92$ METEOR pp. In addition, both results have reported a $p$-value $<0.01$. These results give evidence to both the improved performance provided by the RELAX gradient estimator and the usefulness of the coverage term. In Appendix \ref{subsec:appendix-qual}, we also provide a qualitative example which shows that the increase in METEOR score is most likely given by the positive impact of the coverage term, which has allowed the model to retrieve relevant phrases from the input documents. 


\textbf{WCEP}.
Table \ref{wcep_results} shows the results over the WCEP test set. The trend is similar to that over Multi-News, but the improvements with the proposed models have been even more pronounced. In the first place, the NLL baseline has set a very strong performance compared to the previous work, showing the full potential of a long-input model such as the Longformer for MDS. As for Multi-News, the best results have been achieved with the RELAX gradient estimator, with improvements of up to $+1.32$ ROUGE-1 pp and $+3.17$ METEOR pp over the NLL baseline. The inclusion of the coverage term with RELAX has not been able to increase the ROUGE scores, but has increased METEOR by $+1.64$ pp. Again, we attribute this to the model's improved coverage of the input documents, which leads to an increased number of matches under METEOR's more relaxed matching scheme. A qualitative example is discussed in Appendix \ref{subsec:appendix-qual}. 


\section{Analysis}
\label{sec:analysis}

In this section, we present a more detailed analysis of the impact of the coverage term, the few-shot fine-tuning, and the RELAX gradient estimator using the Multi-News validation set as reference. For a further insight into the coverage reward, we also include an analysis of its trajectory during training. All the selected hyperparameters are listed in Appendix \ref{subsec:appendix-modelhyp}.

\subsection{Impact of the Coverage Term }
\label{subsec:cov-rouge-imp-analysis}
Our rationale for including a coverage term in the reward is to ensure coverage of the input documents beyond what can be driven by the reference summaries alone. We note that this may or may not translate into an improvement of the evaluation metrics, but it seems to add intrinsic value to the summaries nevertheless. For this reason, we further analyze the impact of the coverage term hereafter.

\begin{figure}[!ht]
	\centering
	\includegraphics[width=\linewidth]{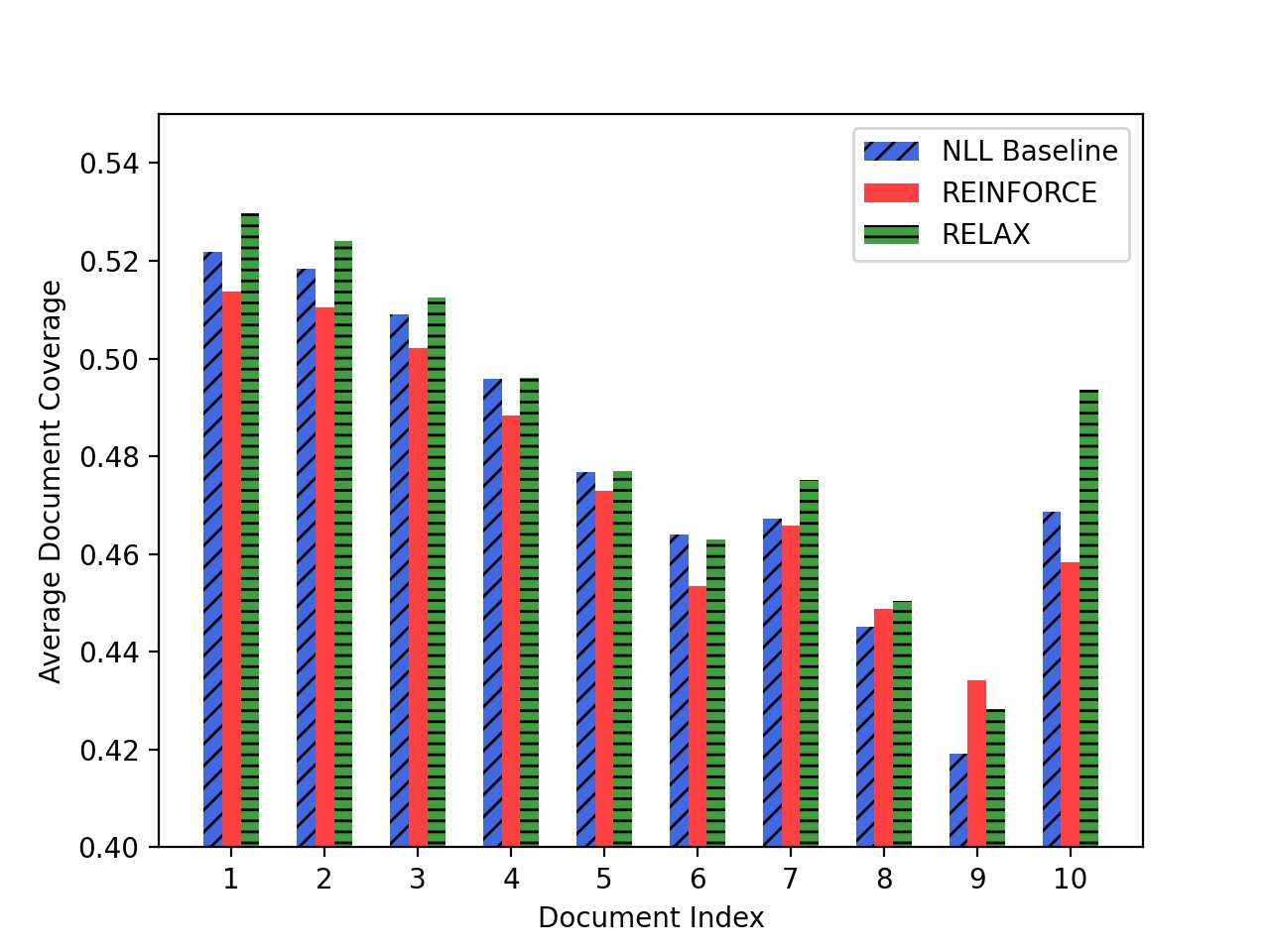}
	\caption{\label{fig:cov-comparison} Comparison of the EFC coverage across the input documents over the Multi-News validation set for the NLL baseline, REINFORCE and RELAX.}
\end{figure}

Figure \ref{fig:cov-comparison} shows the average EFC coverage (\ref{eq:coverage}) for the documents in the input sets, indexed by the document position in the set (first, second etc). The figure shows that the inclusion of the coverage term with RELAX has led to a marked increase of the coverage, and almost evenly distributed across all the documents in the input set. In particular, the document in the last position has achieved the largest coverage improvement.


\begin{figure}[!ht]
	\centering
	\includegraphics[width=\linewidth]{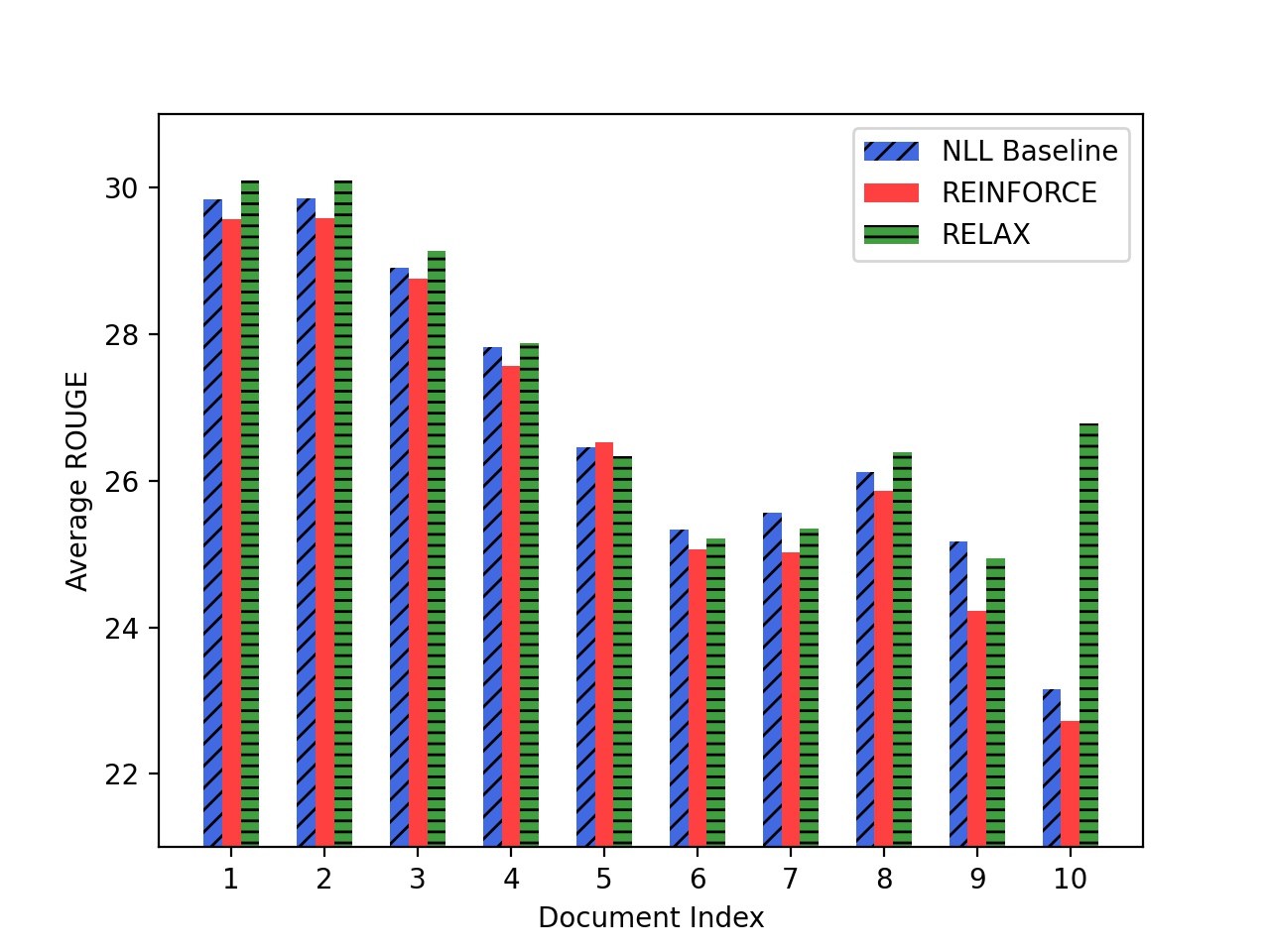}
	\caption{\label{fig:rouge-comparison} Comparison of the average ROUGE score across the input documents over the Multi-News validation set for the NLL baseline, REINFORCE and RELAX. The average is taken over the ROUGE-1, ROUGE-2, and ROUGE-L scores.}
\end{figure}

In turn, Figure \ref{fig:rouge-comparison} shows the average ROUGE score for the documents in the input sets, obtained by averaging the ROUGE-1, ROUGE-2, and ROUGE-L scores computed between the predicted summary and the document (NB: not the reference summary). The figure
shows that the improvements in ROUGE score across the document set are similar to those in EFC coverage, rather evenly distributed, and with an improvement of over $+4$ pp for the document in the last position. This is further evidence that the normalized coverage reward (\ref{eq:final coverage_reward}) is able to drive the model towards predictions that cover the input set more uniformly.


\subsection{Few-Shot Fine-Tuning}
\label{subsec:low-resource-analysis}


To explore the behavior of the few-shot fine-tuning, we compare the validation-set performance on Multi-News with varying number of training examples, from $10$ to $2000$. The model's configuration is the best, with RELAX and the coverage term in the reward.
Table \ref{tab:low-resource} shows that the performance is the highest with 1000 examples, and starts to drop beyond this number. This is an important observation, as it shows that the reinforcement learning objective may lead to undesirable parameterizations beyond a point, and that the number of fine-tuning samples has to be treated as a hyperparameter.


\begin{table}[!ht]
\centering
\begin{tabular}{ccc}
\hline
\textbf{\# Examples} & \textbf{Avg. ROUGE} & \textbf{Avg. MS} \\
\hline
Baseline & 29.84 & 29.48 \\
10 & 29.84 & 29.48 \\
100 & 29.89 & 29.44 \\
1000 & \textbf{30.09} & \textbf{30.04} \\
2000 & 29.80 & 29.39 \\
\hline
\end{tabular}
\caption{Comparison of the average ROUGE and METEOR scores with different fine-tuning sizes over the Multi-News validation set.}
\label{tab:low-resource}
\end{table}

\subsection{Configuring RELAX}
\label{subsec:relax-analysis}

The RELAX gradient estimator introduces two new hyperparameters: the temperature parameter, $\tau$, and the control variate, $c_\phi$. Hereafter, we discuss their impact and design.\\

\begin{figure}[!ht]
	\centering
	\includegraphics[width=\linewidth]{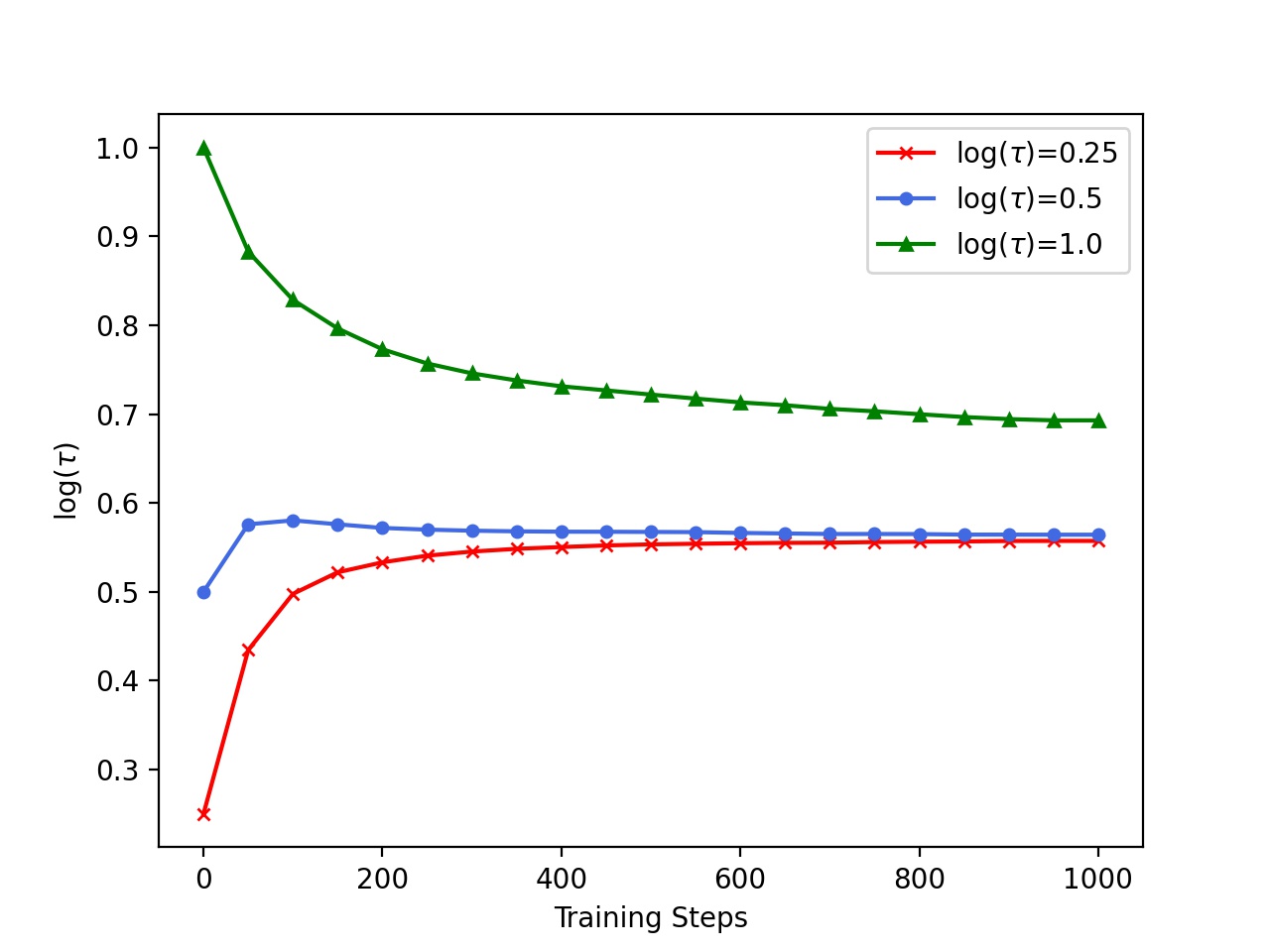}
	\caption{\label{fig:tau-init-change} Trajectory of the log($\tau$) temperature parameter over 1000 Multi-News training steps for different initializations.}
\end{figure}

\textbf{Temperature parameter}.
The RELAX gradient estimator uses a temperature parameter, $\tau$, in the Gumbel-Softmax sampling (\ref{eq:Gumbel-Softmax}). This parameter is maintained in log scale for convenience and is learnable alongside all other parameters; yet, its initial value can have a significant impact on the final model. To explore its behavior, Figure \ref{fig:tau-init-change} shows the trajectory of parameter $\log \tau$ over 1000 Multi-News training steps for different initializations (0.25, 0.5 and 1.0). The trajectories show that, irrespective of its initial value, $\log \tau$ converges to a stable value within approximately 400 training steps. For the initializations at 0.25 and 1.0, within the first 200-300 training steps $\log \tau$ drifts significantly ($\approx \pm 0.25$ units) from its initial value. Conversely, with the intermediate initialization at 0.5, the value remains substantially stable over the whole trajectory.
Since limiting drift during fine-tuning is generally desirable, we have initialized $\log \tau$ to 0.5 in all experiments.


\vspace{12pt}

\textbf{Control variate size}.
\label{subsec:control-variate-analysis}
Many different architectures could be used for the control variate, but given our choice described in Section \ref{subsec:rl_grad_est}, the main parameter is the feed-forward layers' hidden size. To explore its impact, Table \ref{tab:explore_cv} shows the average values of the ROUGE score and the coverage score over the Multi-News validation set with different hidden sizes (128, 256, and 512). The ROUGE score is computed between the prediction and the reference and is the average of ROUGE-1/2/L, while the coverage score is the average EFC of all the input documents. The values in Table \ref{tab:explore_cv} show that, the larger the control variate, the more the model is able to increase the coverage score. However, the average ROUGE score drops beyond a size of 256. 
We speculate that this behavior is due to the larger scale of the coverage reward, as by providing more capacity to the network, we allow the control variate to increasingly correlate with the multi-document coverage reward rather than the ROUGE reward. 
To strike a satisfactory trade-off, we have therefore chosen 256 as the hidden size for all experiments with Multi-News, and carried out an equivalent selection for WCEP.


\begin{table}[!ht]
\centering
\begin{tabular}{ccc}
\hline
\textbf{Hidden Size} & \textbf{Avg. ROUGE} & \textbf{Avg. Coverage} \\
\hline
128 & 30.01 & 0.4821 \\
256 & \textbf{30.09} & 0.4849 \\
512 & 29.91 & \textbf{0.5038} \\
\hline
\end{tabular}
\caption{Comparison of the average ROUGE and coverage scores over the Multi-News validation set with different hidden sizes of the control variate.}
\label{tab:explore_cv}
\end{table}

\subsection{Coverage Reward Trajectory}
\label{subsec:reward-analysis}
\begin{figure}[!ht]
	\centering
	\includegraphics[width=\linewidth]{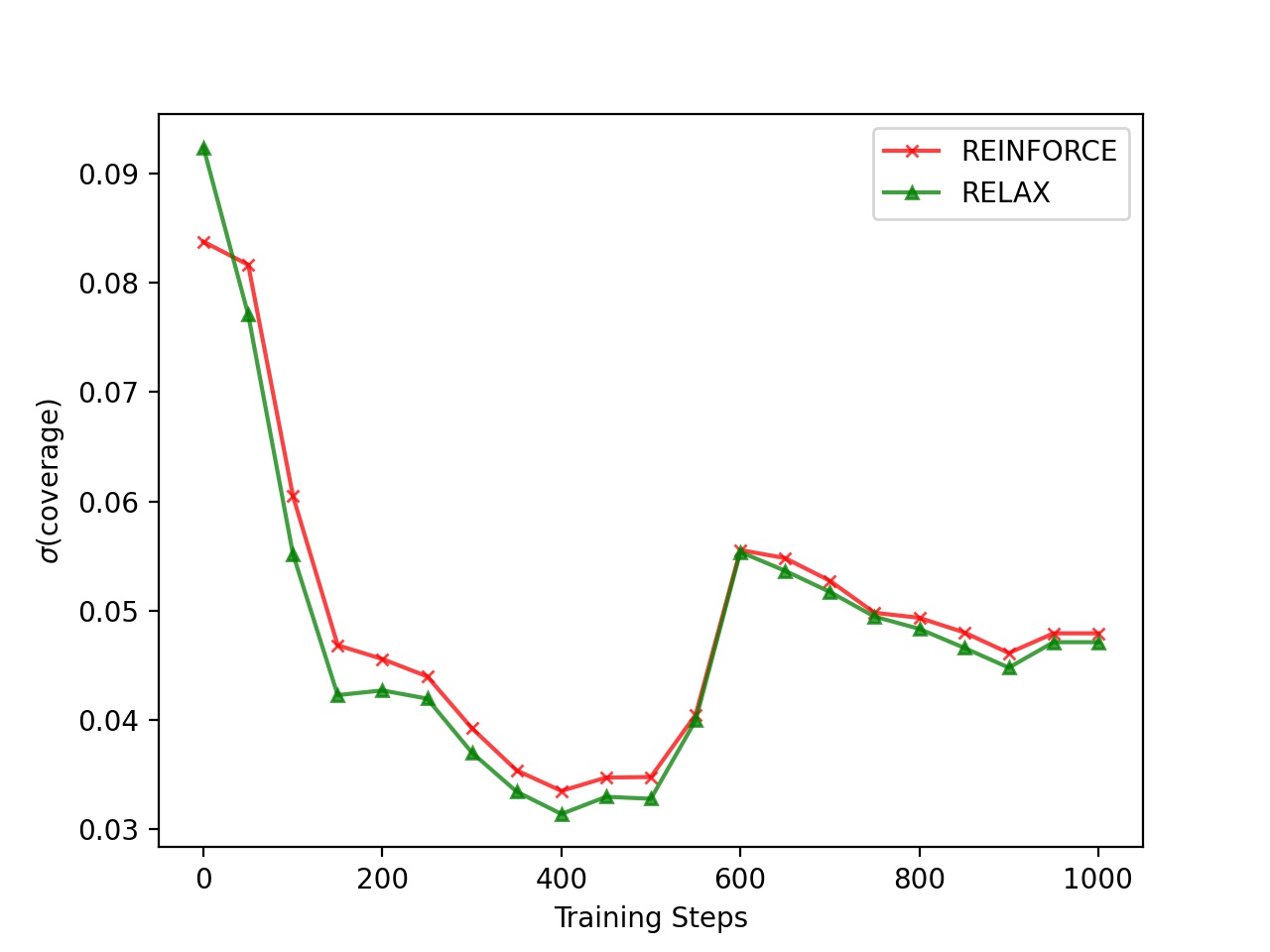}
	\caption{\label{fig:std-cov} Standard deviation of the coverage scores (EFC) across the input documents for REINFORCE and RELAX against the training step. For both estimators, the standard deviation drops below $0.06$ very early into the training, and sets to approximately $0.05$.}
\end{figure}

In a reinforcement learning framework, it could be useful to monitor the value of the reward over the training steps. Typically, the reward should exhibit an upward trajectory, since the reward should tend to increase as the model learns to make better predictions. In our case, we look to explore the impact of our coverage reward on the coverage distribution over the input documents. In particular, we want to verify whether the coverage reward is able to promote predictions that cover the input documents more evenly, which should translate into a decreased standard deviation. To this aim, Figure \ref{fig:std-cov} shows a plot of the standard deviation of the coverage scores (EFC) across the input document set against the training step. The trajectories show that both REINFORCE and RELAX have been able to decrease the standard deviation of the predictions to approximately $0.05$ units from initial values of $0.08-0.09$. The drop in standard deviation occurs quite quickly during training, coinciding with the improvement in the reward value of the predictions. Comparing REINFORCE with RELAX also shows that RELAX has been able to achieve lower standard deviation values throughout the training, with the exception of the very start.

\section{Conclusion}
\label{sec:conclusion}

In this paper, we have proposed fine-tuning a multi-document summarization model with a reward that balances the use of the reference summaries with the coverage of the input documents within a reinforcement learning framework. The rationale for the proposed reward is that the reference summaries alone may not be sufficient for an effective fine-tuning of the model in the presence of very large inputs such as those typical of MDS datasets. Another key component of the proposed approach is the use of a modern gradient estimator of the policy gradient, RELAX. The experimental results over two news-based MDS datasets, Multi-News and WCEP, have shown that the proposed approach has been able to achieve a marked improvement of ROUGE and METEOR scores compared to its NLL-pretrained baseline, and prove competitive against most existing approaches. In addition, the proposed approach has been able to increase the coverage of the input documents, and evenly across the entire document set. As future work, we aim to explore ways to prevent or mollify the model's drift with larger number of training steps, and explore alternative architectures and configurations for the control variate of the RELAX estimator.

\bibliography{anthology,custom}

\begin{thebibliography}{42}
\expandafter\ifx\csname natexlab\endcsname\relax\def\natexlab#1{#1}\fi

\bibitem[{Beltagy et~al.(2020)Beltagy, Peters, and
  Cohan}]{Beltagy2020Longformer}
Iz~Beltagy, Matthew~E. Peters, and Arman Cohan. 2020.
\newblock Longformer: The long-document transformer.
\newblock \emph{arXiv:2004.05150}.

\bibitem[{Bengio et~al.(2015)Bengio, Vinyals, Jaitly, and Shazeer}]{bengio2015}
Samy Bengio, Oriol Vinyals, Navdeep Jaitly, and Noam Shazeer. 2015.
\newblock Scheduled sampling for sequence prediction with recurrent neural
  networks.
\newblock In \emph{Proceedings of the 28th International Conference on Neural
  Information Processing Systems - Volume 1}, NIPS'15, page 1171–1179,
  Cambridge, MA, USA. MIT Press.

\bibitem[{Carbonell and Goldstein(1998)}]{carbonell1998}
Jaime Carbonell and Jade Goldstein. 1998.
\newblock \href {https://doi.org/10.1145/290941.291025} {The use of mmr,
  diversity-based reranking for reordering documents and producing summaries}.
\newblock In \emph{Proceedings of the 21st Annual International ACM SIGIR
  Conference on Research and Development in Information Retrieval}, SIGIR
  ’98, page 335–336, New York, NY, USA. Association for Computing
  Machinery.

\bibitem[{Christensen et~al.(2013)Christensen, {Mausam}, Soderland, and
  Etzioni}]{christensen-etal-2013-towards}
Janara Christensen, {Mausam}, Stephen Soderland, and Oren Etzioni. 2013.
\newblock \href {https://aclanthology.org/N13-1136} {Towards coherent
  multi-document summarization}.
\newblock In \emph{Proceedings of the 2013 Conference of the North {A}merican
  Chapter of the Association for Computational Linguistics: Human Language
  Technologies}, pages 1163--1173, Atlanta, Georgia. Association for
  Computational Linguistics.

\bibitem[{Devlin et~al.(2019)Devlin, Chang, Lee, and Toutanova}]{devlin2018}
Jacob Devlin, Ming-Wei Chang, Kenton Lee, and Kristina Toutanova. 2019.
\newblock \href {https://doi.org/10.18653/v1/N19-1423} {{BERT}: Pre-training of
  deep bidirectional transformers for language understanding}.
\newblock In \emph{Proceedings of the 2019 Conference of the North {A}merican
  Chapter of the Association for Computational Linguistics: Human Language
  Technologies, Volume 1 (Long and Short Papers)}, pages 4171--4186,
  Minneapolis, Minnesota. Association for Computational Linguistics.

\bibitem[{Ding and Soricut(2017)}]{ding2017coldrl}
Nan Ding and Radu Soricut. 2017.
\newblock \href
  {https://proceedings.neurips.cc/paper/2017/file/faafda66202d234463057972460c04f5-Paper.pdf}
  {Cold-start reinforcement learning with softmax policy gradient}.
\newblock In \emph{Advances in Neural Information Processing Systems},
  volume~30. Curran Associates, Inc.

\bibitem[{Dreyer et~al.(2021)Dreyer, Liu, Nan, Atluri, and Ravi}]{dreyer2021}
Markus Dreyer, Mengwen Liu, Feng Nan, Sandeep Atluri, and Sujith Ravi. 2021.
\newblock \href {http://arxiv.org/abs/2108.02859} {Analyzing the
  abstractiveness-factuality tradeoff with nonlinear abstractiveness
  constraints}.
\newblock \emph{CoRR}, abs/2108.02859.

\bibitem[{Dror et~al.(2018)Dror, Baumer, Shlomov, and Reichart}]{dror2018}
Rotem Dror, Gili Baumer, Segev Shlomov, and Roi Reichart. 2018.
\newblock \href {http://aclweb.org/anthology/P18-1128} {The hitchhiker's guide
  to testing statistical significance in natural language processing}.
\newblock In \emph{Proceedings of the 56th Annual Meeting of the Association
  for Computational Linguistics (Volume 1: Long Papers)}, pages 1383--1392.
  Association for Computational Linguistics.

\bibitem[{Erkan and Radev(2004)}]{erkan-and-radev04}
G\"{u}nes Erkan and Dragomir~R. Radev. 2004.
\newblock Lexrank: Graph-based lexical centrality as salience in text
  summarization.
\newblock \emph{Journal of Artificial Intelligence Research}, 22:457–479.

\bibitem[{Fabbri et~al.(2019)Fabbri, Li, She, Li, and Radev}]{fabbri2019}
Alexander Fabbri, Irene Li, Tianwei She, Suyi Li, and Dragomir Radev. 2019.
\newblock \href {https://doi.org/10.18653/v1/P19-1102} {Multi-news: A
  large-scale multi-document summarization dataset and abstractive hierarchical
  model}.
\newblock In \emph{Proceedings of the 57th Annual Meeting of the Association
  for Computational Linguistics}, pages 1074--1084, Florence, Italy.
  Association for Computational Linguistics.

\bibitem[{Fabbri et~al.(2021)Fabbri, Kryściński, McCann, Xiong, Socher, and
  Radev}]{summeval2021}
Alexander~R. Fabbri, Wojciech Kryściński, Bryan McCann, Caiming Xiong,
  Richard Socher, and Dragomir Radev. 2021.
\newblock \href {https://doi.org/10.1162/tacl_a_00373} {{SummEval:
  Re-evaluating Summarization Evaluation}}.
\newblock \emph{Transactions of the Association for Computational Linguistics},
  9:391--409.

\bibitem[{Gholipour~Ghalandari et~al.(2020)Gholipour~Ghalandari, Hokamp, Pham,
  Glover, and Ifrim}]{gholipour-ghalandari-etal-2020-large}
Demian Gholipour~Ghalandari, Chris Hokamp, Nghia~The Pham, John Glover, and
  Georgiana Ifrim. 2020.
\newblock \href {https://doi.org/10.18653/v1/2020.acl-main.120} {A large-scale
  multi-document summarization dataset from the {W}ikipedia current events
  portal}.
\newblock In \emph{Proceedings of the 58th Annual Meeting of the Association
  for Computational Linguistics}, pages 1302--1308, Online. Association for
  Computational Linguistics.

\bibitem[{Grathwohl et~al.(2018)Grathwohl, Choi, Wu, Roeder, and
  Duvenaud}]{grathwohl2018backpropagation}
Will Grathwohl, Dami Choi, Yuhuai Wu, Geoff Roeder, and David Duvenaud. 2018.
\newblock \href {https://openreview.net/forum?id=SyzKd1bCW} {Backpropagation
  through the void: Optimizing control variates for black-box gradient
  estimation}.
\newblock In \emph{International Conference on Learning Representations}.

\bibitem[{Grusky et~al.(2018)Grusky, Naaman, and Artzi}]{grusky2018newsroom}
Max Grusky, Mor Naaman, and Yoav Artzi. 2018.
\newblock \href {https://doi.org/10.18653/v1/N18-1065} {{N}ewsroom: A dataset
  of 1.3 million summaries with diverse extractive strategies}.
\newblock In \emph{Proceedings of the 2018 Conference of the North {A}merican
  Chapter of the Association for Computational Linguistics: Human Language
  Technologies, Volume 1 (Long Papers)}, pages 708--719, New Orleans,
  Louisiana. Association for Computational Linguistics.

\bibitem[{Hokamp et~al.(2020)Hokamp, Ghalandari, Pham, and Glover}]{hokamp2020}
Chris Hokamp, Demian~Gholipour Ghalandari, Nghia~The Pham, and John Glover.
  2020.
\newblock \href {http://arxiv.org/abs/2006.08748} {Dyne: Dynamic ensemble
  decoding for multi-document summarization}.
\newblock \emph{CoRR}, abs/2006.08748.

\bibitem[{Jang et~al.(2017)Jang, Gu, and Poole}]{jang2017categorical}
Eric Jang, Shixiang Gu, and Ben Poole. 2017.
\newblock \href {http://arxiv.org/abs/1611.01144} {Categorical
  reparameterization with gumbel-softmax}.
\newblock In \emph{International Conference on Learning Representations}.

\bibitem[{Kry{\'s}ci{\'n}ski et~al.(2018)Kry{\'s}ci{\'n}ski, Paulus, Xiong, and
  Socher}]{kryscinski2018abstraction}
Wojciech Kry{\'s}ci{\'n}ski, Romain Paulus, Caiming Xiong, and Richard Socher.
  2018.
\newblock \href {https://doi.org/10.18653/v1/D18-1207} {Improving abstraction
  in text summarization}.
\newblock In \emph{Proceedings of the 2018 Conference on Empirical Methods in
  Natural Language Processing}, pages 1808--1817, Brussels, Belgium.
  Association for Computational Linguistics.

\bibitem[{Lavie and Agarwal(2007)}]{lavie-agarwal-2007-meteor}
Alon Lavie and Abhaya Agarwal. 2007.
\newblock \href {https://aclanthology.org/W07-0734} {{METEOR}: An automatic
  metric for {MT} evaluation with high levels of correlation with human
  judgments}.
\newblock In \emph{Proceedings of the Second Workshop on Statistical Machine
  Translation}, pages 228--231, Prague, Czech Republic. Association for
  Computational Linguistics.

\bibitem[{Lebanoff et~al.(2018)Lebanoff, Song, and Liu}]{lebanoff2018}
Logan Lebanoff, Kaiqiang Song, and Fei Liu. 2018.
\newblock \href {https://doi.org/10.18653/v1/D18-1446} {Adapting the neural
  encoder-decoder framework from single to multi-document summarization}.
\newblock In \emph{Proceedings of the 2018 Conference on Empirical Methods in
  Natural Language Processing}, pages 4131--4141, Brussels, Belgium.
  Association for Computational Linguistics.

\bibitem[{Lewis et~al.(2020)Lewis, Liu, Goyal, Ghazvininejad, Mohamed, Levy,
  Stoyanov, and Zettlemoyer}]{lewis2020}
Mike Lewis, Yinhan Liu, Naman Goyal, Marjan Ghazvininejad, Abdelrahman Mohamed,
  Omer Levy, Veselin Stoyanov, and Luke Zettlemoyer. 2020.
\newblock \href {https://doi.org/10.18653/v1/2020.acl-main.703} {{BART}:
  Denoising sequence-to-sequence pre-training for natural language generation,
  translation, and comprehension}.
\newblock In \emph{Proceedings of the 58th Annual Meeting of the Association
  for Computational Linguistics}, pages 7871--7880, Online. Association for
  Computational Linguistics.

\bibitem[{Li et~al.(2019)Li, Lei, Qin, and Wang}]{li2019dsr}
Siyao Li, Deren Lei, Pengda Qin, and William~Yang Wang. 2019.
\newblock \href {https://doi.org/10.18653/v1/D19-1623} {Deep reinforcement
  learning with distributional semantic rewards for abstractive summarization}.
\newblock In \emph{Proceedings of the 2019 Conference on Empirical Methods in
  Natural Language Processing and the 9th International Joint Conference on
  Natural Language Processing (EMNLP-IJCNLP)}, pages 6038--6044, Hong Kong,
  China. Association for Computational Linguistics.

\bibitem[{Li et~al.(2020)Li, Xiao, Liu, Wu, Wang, and
  Du}]{li-etal-2020-leveraging-graph}
Wei Li, Xinyan Xiao, Jiachen Liu, Hua Wu, Haifeng Wang, and Junping Du. 2020.
\newblock \href {https://doi.org/10.18653/v1/2020.acl-main.555} {Leveraging
  graph to improve abstractive multi-document summarization}.
\newblock In \emph{Proceedings of the 58th Annual Meeting of the Association
  for Computational Linguistics}, pages 6232--6243, Online. Association for
  Computational Linguistics.

\bibitem[{Lin(2004)}]{lin2004}
Chin-Yew Lin. 2004.
\newblock \href {https://www.aclweb.org/anthology/W04-1013} {{ROUGE}: A package
  for automatic evaluation of summaries}.
\newblock In \emph{Text Summarization Branches Out}, pages 74--81, Barcelona,
  Spain. Association for Computational Linguistics.

\bibitem[{Liu et~al.(2018)Liu, Saleh, Pot, Goodrich, Sepassi, Kaiser, and
  Shazeer}]{j.2018generating}
Peter~J. Liu, Mohammad Saleh, Etienne Pot, Ben Goodrich, Ryan Sepassi, Lukasz
  Kaiser, and Noam Shazeer. 2018.
\newblock \href {https://openreview.net/forum?id=Hyg0vbWC-} {Generating
  wikipedia by summarizing long sequences}.
\newblock In \emph{International Conference on Learning Representations}.

\bibitem[{Liu and Lapata(2019)}]{liu-lapata-2019-hierarchical}
Yang Liu and Mirella Lapata. 2019.
\newblock \href {https://doi.org/10.18653/v1/P19-1500} {Hierarchical
  transformers for multi-document summarization}.
\newblock In \emph{Proceedings of the 57th Annual Meeting of the Association
  for Computational Linguistics}, pages 5070--5081, Florence, Italy.
  Association for Computational Linguistics.

\bibitem[{Mani and Bloedorn(1997)}]{mani-bloedorn-97}
Inderjeet Mani and Eric Bloedorn. 1997.
\newblock Multi-document summarization by graph search and matching.
\newblock In \emph{Proceedings of the Fourteenth National Conference on
  Artificial Intelligence and Ninth Conference on Innovative Applications of
  Artificial Intelligence}, AAAI'97/IAAI'97, page 622–628. AAAI Press.

\bibitem[{Mao et~al.(2020)Mao, Qu, Xie, Ren, and Han}]{mao-etal-2020-multi}
Yuning Mao, Yanru Qu, Yiqing Xie, Xiang Ren, and Jiawei Han. 2020.
\newblock \href {https://doi.org/10.18653/v1/2020.emnlp-main.136}
  {Multi-document summarization with maximal marginal relevance-guided
  reinforcement learning}.
\newblock In \emph{Proceedings of the 2020 Conference on Empirical Methods in
  Natural Language Processing (EMNLP)}, pages 1737--1751, Online. Association
  for Computational Linguistics.

\bibitem[{Nallapati et~al.(2017)Nallapati, Zhai, and Zhou}]{nallapati2017}
Ramesh Nallapati, Feifei Zhai, and Bowen Zhou. 2017.
\newblock Summarunner: A recurrent neural network based sequence model for
  extractive summarization of documents.
\newblock In \emph{Proceedings of the Thirty-First AAAI Conference on
  Artificial Intelligence}, AAAI'17, page 3075–3081. AAAI Press.

\bibitem[{Nayeem et~al.(2018)Nayeem, Fuad, and
  Chali}]{nayeem-etal-2018-abstractive}
Mir~Tafseer Nayeem, Tanvir~Ahmed Fuad, and Yllias Chali. 2018.
\newblock \href {https://aclanthology.org/C18-1102} {Abstractive unsupervised
  multi-document summarization using paraphrastic sentence fusion}.
\newblock In \emph{Proceedings of the 27th International Conference on
  Computational Linguistics}, pages 1191--1204, Santa Fe, New Mexico, USA.
  Association for Computational Linguistics.

\bibitem[{Parnell et~al.(2021)Parnell, Jauregi~Unanue, and
  Piccardi}]{parnell-etal-2021-rewardsofsum}
Jacob Parnell, Inigo Jauregi~Unanue, and Massimo Piccardi. 2021.
\newblock \href {https://doi.org/10.18653/v1/2021.spnlp-1.1}
  {{R}ewards{O}f{S}um: Exploring reinforcement learning rewards for
  summarisation}.
\newblock In \emph{Proceedings of the 5th Workshop on Structured Prediction for
  NLP (SPNLP 2021)}, pages 1--11, Online. Association for Computational
  Linguistics.

\bibitem[{Pasunuru et~al.(2021)Pasunuru, Liu, Bansal, Ravi, and
  Dreyer}]{pasunuru-etal-2021-efficiently}
Ramakanth Pasunuru, Mengwen Liu, Mohit Bansal, Sujith Ravi, and Markus Dreyer.
  2021.
\newblock \href {https://doi.org/10.18653/v1/2021.naacl-main.380} {Efficiently
  summarizing text and graph encodings of multi-document clusters}.
\newblock In \emph{Proceedings of the 2021 Conference of the North American
  Chapter of the Association for Computational Linguistics: Human Language
  Technologies}, pages 4768--4779, Online. Association for Computational
  Linguistics.

\bibitem[{Paulus et~al.(2018)Paulus, Xiong, and Socher}]{paulus2017deep}
Romain Paulus, Caiming Xiong, and Richard Socher. 2018.
\newblock \href {https://openreview.net/forum?id=HkAClQgA-} {A deep reinforced
  model for abstractive summarization}.
\newblock In \emph{International Conference on Learning Representations}.

\bibitem[{Ranzato et~al.(2016)Ranzato, Chopra, Auli, and
  Zaremba}]{ranzato2016sequence}
Marc'Aurelio Ranzato, Sumit Chopra, Michael Auli, and Wojciech Zaremba. 2016.
\newblock \href {http://arxiv.org/abs/1511.06732} {Sequence level training with
  recurrent neural networks}.
\newblock In \emph{4th International Conference on Learning Representations,
  {ICLR} 2016, San Juan, Puerto Rico, May 2-4, 2016, Conference Track
  Proceedings}.

\bibitem[{{Rennie} et~al.(2017){Rennie}, {Marcheret}, {Mroueh}, {Ross}, and
  {Goel}}]{rennie2017selfcritical}
S.~J. {Rennie}, E.~{Marcheret}, Y.~{Mroueh}, J.~{Ross}, and V.~{Goel}. 2017.
\newblock \href {https://doi.org/10.1109/CVPR.2017.131} {Self-critical sequence
  training for image captioning}.
\newblock In \emph{2017 IEEE Conference on Computer Vision and Pattern
  Recognition (CVPR)}, pages 1179--1195.

\bibitem[{Rush et~al.(2015)Rush, Chopra, and Weston}]{rush2015}
Alexander~M. Rush, Sumit Chopra, and Jason Weston. 2015.
\newblock \href {https://doi.org/10.18653/v1/D15-1044} {A neural attention
  model for abstractive sentence summarization}.
\newblock In \emph{Proceedings of the 2015 Conference on Empirical Methods in
  Natural Language Processing}, pages 379--389, Lisbon, Portugal. Association
  for Computational Linguistics.

\bibitem[{See et~al.(2017)See, Liu, and Manning}]{see-etal-2017-get}
Abigail See, Peter~J. Liu, and Christopher~D. Manning. 2017.
\newblock \href {https://doi.org/10.18653/v1/P17-1099} {Get to the point:
  Summarization with pointer-generator networks}.
\newblock In \emph{Proceedings of the 55th Annual Meeting of the Association
  for Computational Linguistics (Volume 1: Long Papers)}, pages 1073--1083,
  Vancouver, Canada. Association for Computational Linguistics.

\bibitem[{Sutton et~al.(1999)Sutton, McAllester, Singh, and
  Mansour}]{sutton1999policy}
Richard~S. Sutton, David McAllester, Satinder Singh, and Yishay Mansour. 1999.
\newblock Policy gradient methods for reinforcement learning with function
  approximation.
\newblock In \emph{Proceedings of the 12th International Conference on Neural
  Information Processing Systems}, NIPS'99, page 1057–1063, Cambridge, MA,
  USA. MIT Press.

\bibitem[{Williams(1992)}]{williams92reinforce}
Ronald~J. Williams. 1992.
\newblock \href {https://doi.org/10.1007/BF00992696} {Simple statistical
  gradient-following algorithms for connectionist reinforcement learning}.
\newblock \emph{Mach. Learn.}, 8(3–4):229–256.

\bibitem[{Zaheer et~al.(2020)Zaheer, Guruganesh, Dubey, Ainslie, Alberti,
  Ontanon, Pham, Ravula, Wang, Yang, and Ahmed}]{zaheerbigbird2020}
Manzil Zaheer, Guru Guruganesh, Kumar~Avinava Dubey, Joshua Ainslie, Chris
  Alberti, Santiago Ontanon, Philip Pham, Anirudh Ravula, Qifan Wang, Li~Yang,
  and Amr Ahmed. 2020.
\newblock \href
  {https://proceedings.neurips.cc/paper/2020/file/c8512d142a2d849725f31a9a7a361ab9-Paper.pdf}
  {Big bird: Transformers for longer sequences}.
\newblock In \emph{Advances in Neural Information Processing Systems},
  volume~33, pages 17283--17297. Curran Associates, Inc.

\bibitem[{Zhang et~al.(2018)Zhang, Tan, and Wan}]{zhang-etal-2018-adapting}
Jianmin Zhang, Jiwei Tan, and Xiaojun Wan. 2018.
\newblock \href {https://doi.org/10.18653/v1/W18-6545} {Adapting neural
  single-document summarization model for abstractive multi-document
  summarization: A pilot study}.
\newblock In \emph{Proceedings of the 11th International Conference on Natural
  Language Generation}, pages 381--390, Tilburg University, The Netherlands.
  Association for Computational Linguistics.

\bibitem[{Zhang et~al.(2020{\natexlab{a}})Zhang, Zhao, Saleh, and
  Liu}]{zhang2020}
Jingqing Zhang, Yao Zhao, Mohammad Saleh, and Peter Liu. 2020{\natexlab{a}}.
\newblock \href {http://proceedings.mlr.press/v119/zhang20ae.html} {{PEGASUS}:
  Pre-training with extracted gap-sentences for abstractive summarization}.
\newblock In \emph{Proceedings of the 37th International Conference on Machine
  Learning}, volume 119 of \emph{Proceedings of Machine Learning Research},
  pages 11328--11339. PMLR.

\bibitem[{Zhang et~al.(2020{\natexlab{b}})Zhang, Kishore, Wu, Weinberger, and
  Artzi}]{zhang2020bertscore}
Tianyi Zhang, Varsha Kishore, Felix Wu, Kilian~Q. Weinberger, and Yoav Artzi.
  2020{\natexlab{b}}.
\newblock \href {http://arxiv.org/abs/1904.09675} {Bertscore: Evaluating text
  generation with bert}.
\newblock In \emph{International Conference on Learning Representations}.

\end{thebibliography}
\bibliographystyle{acl_natbib}

\newpage
\appendix

\section{Model and Datasets}
\label{sec:appendix}

\subsection{Model Hyperparameters}
\label{subsec:appendix-modelhyp}

Our baseline model is the BART Longformer encoder-decoder (BART-LED) of \citet{Beltagy2020Longformer}. In all experiments, it has been first pre-trained over the training set using the negative log-likelihood until convergence, which has typically set within 5 epochs, in line with what reported in \citet{pasunuru-etal-2021-efficiently}. The best models on the validation set (average ROUGE) have then been fine-tuned with the reinforcement learning objectives in (\ref{eq:reinforce}) and (\ref{eq:relax}). BART-LED has $459$M parameters, and the addition of the control variate for the RELAX experiments adds approximately $12$M more parameters. We have used the Adam optimizer for training both the main BART-LED model and the parameters of the control variate of the RELAX gradient estimator. The learning rate of the main optimizer for the pre-training of the baseline has been set to the default of $3\times 10^{-5}$ \cite{Beltagy2020Longformer}, before being reduced to $3\times 10^{-6}$ for fine-tuning with the reinforcement learning approaches. For training the control variate, we have set an initial learning rate of $1\times 10^{-2}$ and initialized the learnable $\log \tau$ parameter with a value of $0.5$. The optimal size of the hidden layers of the control variate appears to empirically correlate with the maximum length of the reference summaries in the dataset. For Multi-News, we have set the size to be $256$, and for WCEP to $40$\footnote{\citet{gholipour-ghalandari-etal-2020-large} indicate that the annotation guidelines for WCEP suggested $40$ as maximum summary length.}. For the multi-document coverage reward, we have used a $\beta$ value of $1.0$. 
The entire model has been implemented on top of PyTorch Lightning\footnote{\url{https://github.com/PyTorchLightning/pytorch-lightning}}. Please refer to Table \ref{tab:hyperparameters} for a full list of the hyperparameters. For all experiments, we have used an NVIDIA Quadro RTX 6000 with 24 GB of memory.

\begin{table}[!ht]
\centering
\begin{tabular}{ccc}
\hline
\textbf{Hyperparameter} & \textbf{Multi-News} & \textbf{WCEP} \\
\hline
Learning Rate (Train) & $3\times 10^{-5}$ & $3\times 10^{-5}$ \\
Learning Rate (Tune) & $3\times 10^{-6}$ & $3\times 10^{-6}$ \\
Learning Rate (RELAX) & $1\times 10^{-2}$ & $1\times 10^{-2}$ \\
$\log(\tau)$ (RELAX) & 0.5 & 0.5 \\
Hidden Size (RELAX) & 256 & 40 \\
$\beta$ (Coverage) & 1.0 & 1.0 \\
Max Input Length & 16384 & 16384 \\
Max Output Length & 256 & 40 \\
Label Smoothing & 0.0 & 0.0 \\
Training Epochs & 5 & 5 \\
Tuning Epochs & 1 & 1 \\
Batch Size & 1 & 1 \\
Beam Size & 1 & 1 \\
\hline
\end{tabular}
\caption{Hyperparameters used for training and evaluation.}
\label{tab:hyperparameters}
\end{table}


\subsection{Dataset Links and Statistics}
\textbf{Multi-News}. Accessible via the Hugging Face Datasets Python package: \url{https://github.com/huggingface/datasets/tree/master/datasets/multi_news}.

For fine-tuning, we have pulled the raw data from the authors' own repository: \url{https://github.com/Alex-Fabbri/Multi-News}.

\textbf{WCEP}. Accessible from the following repository: \url{https://github.com/complementizer/wcep-mds-dataset}. \\

Table \ref{tab:dataset_statistics} shows the datasets' main statistics, including the number of samples within each split and the average length in tokens of the reference summaries. For our model, we have used the average length of the reference summaries as a guidance for choosing the maximum output length and the hidden size of the control variate.

\label{subsec:appendix-dataset-stats}
\begin{table}[!ht]
\centering
\begin{tabular}{ccccc}
\hline
\textbf{DS} & \textbf{Train} & \textbf{Test} & \textbf{Dev} & \textbf{Avg. Tokens}\\
\hline
\textbf{M-N} & 44.9K & 5.6K & 5.6K & 263.7 \\
\textbf{WCEP} & 8.1K & 1.0K & 1.0K & 33.3 \\
\hline
\end{tabular}
\caption{Main statistics of the datasets used in the experiments. Multi-News has up to 10 individual articles in each document set, while WCEP has up to 100. The document split sizes have been rounded.}
\label{tab:dataset_statistics}
\end{table}

\vspace{-12pt}

\subsection{Scale of the Coverage Reward}
\label{subsec:appendix-scaling-coverage-term-analysis}

The multi-document reward is used in (\ref{eq:cov_reward+rouge}) in convex combination with the ROUGE-L F1 score, and an appropriate value of the mixing coefficient, $\beta$, needs to be explored. To this aim, Table \ref{tab:explore_beta} shows the values of the average ROUGE and coverage scores over the Multi-News validation set. The coverage has not increased monotonically with the increase of the $\beta$ coefficient. In turn, the ROUGE score has reached a maximum for $\beta = 1.0$. As a trade-off, we have set $\beta = 1.0$ in all experiments. 

\begin{table}[!ht]
\centering
\begin{tabular}{ccc}
\hline
\boldmath{\textbf{$\beta$}} & \textbf{Avg. ROUGE} & \textbf{Avg. Coverage} \\
\hline
0.5 & 29.97 & \textbf{0.5074} \\
1.0 & \textbf{30.09} & 0.4849 \\
2.0 & 30.00 & 0.5068 \\
5.0 & 29.87 & 0.4823 \\
\hline
\end{tabular}
\caption{Average ROUGE and coverage scores over the Multi-News validation set for different values of the reward mixing coefficient, $\beta$.}
\label{tab:explore_beta}
\end{table}

\section{Qualitative Analysis}
\label{subsec:appendix-qual}
Tables \ref{tab:qual_example1} through \ref{tab:qual_example4} present two qualitative examples, one per dataset, where we specifically compare our RELAX implementation with and without the use of the coverage term in the reward. Key points are highlighted in various colors, comments are addressed in the captions, and the ROUGE and METEOR scores are reported for each prediction. Document sets longer than a full page have been truncated to fit.

\begin{table*}[!ht]
\small
    \begin{tabular}{p{1.0\textwidth}}
        \toprule
\textbf{Source Document} \\\midrule
Plenty of churches contain relics of saints, but not many of those relics were found in excavations from sixth-century churches.     Archaeologists at a medieval fortress site in Burgas, Bulgaria, found a lead vessel, which contains some of the ashes from the alleged grave of John the Apostle, in a reliquary that dates to the sixth century C.E. The reliquary, which was once part of an early Christian basilica, is named for Saint John the Theologian, who is considered one of Jesus' apostles. The vessel, which is less than an inch long, is decorated with crosses. Milen Nikolov, director of the Burgas Regional Museum of History, said that early Christians would have believed the relic had healing properties. John the Apostle's grave in Turkey was also a pilgrimage site for early Christians seeking healing, Ancient Origins reports. Nikolov said the reliquary was "one of the most important discoveries" in the museum's history.      In addition to the relic, the archaeologists also uncovered a 10th century Bulgarian royal seal at the fortress site. Meghan DeMaria \texttt{[END]} Ashes from the grave of John the Apostle, one of the Twelve Apostles of Jesus Christ, have been discovered in a lead tube reliquary by Bulgarian archaeologists during excavations of the ancient and medieval port of Burgos (also known as Poros) on Cape Foros in today’s Black Sea city of Burgas.      The discovery of the lead tube containing ashes from the grave of John the Apostle, who is known as St. John the Theologian in Bulgarian (Eastern) Orthodox Christianity, located in the ancient city of Ephesus in Anatolia, today’s Turkey, has been made during the 2014 excavations of the fortress of Burgos (or Poros) on Cape Foros in Burgas but was announced only on Wednesday, March 25, 2015, by Milen Nikolov, Director of the Burgas Regional Museum of History, at a special press conference.      He has also announced other intriguing finds such as the discovery of a Late Antiquity latrine, also found at Burgos (Poros), and the discovery of a 10th century Bulgarian royal seal from the Rusocastro Fortress.      The structures at the ancient and medieval fortress and port of Burgos (Poros) which were excavated in 2014 include an Early Christian basilica from the 6th century AD, a building complex from the 5th-6th century AD, and a Roman villa from the 3rd century AD. The John the Apostle reliquary was found in the 6th century basilica. “Probably a pilgrim from the Foros Peninsula (Cape) went on a pilgrimage to Ephesus, and came back here with this relic which was then donated to the basilica on Foros,” Nikolov has explained, as cited by local news site Gramofona.  Nikolov has described the finding of the reliquary as “one of the most important discoveries in the history of the [Burgas Regional History] Museum”, and the lead tube as a “holy possession that preserved a holy substance” having to do with the beliefs that every year on May 8, the date of John the Apostle’s death, there is manna, a holy curing powder, on the site of his grave.      The lead tube reliquary itself containing the ashes from the grave of John the Apostle (St. John the Theologian) is really tiny: it is only 2.2 cm (less than an inch) long, and its diameter measures 1.7 cm.      The reliquary is dated to the 6th century AD when pilgrimage to the Holy Lands was very common among Christians, Nikolov explains. On one of its sides there is an image of a cross with equal arms inside a medallion, and on the opposite side there is an image of two overlapping crosses with equal arms. The neck of the tube is also decorated with crosses. It has only one handle left, the other has broken off.      In addition to the so called Empty Tomb, i.e. the Tomb of Jesus Christ in Jerusalem, the other centers of Christian pilgrimage in the 6th century AD included the grave of St. Menas in Abu Mina in Egypt; the grave of St. Simeon Stylites the Elder in Antioch (in today’s Turkey); the grave of St. Thecla (or Tecla) in Seleucia, Mesopotamia; the grave of St. Isidore of Chios on the Aegean island of Chios; and the graves of John the Apostle (St. John the Theologian), St. Mary Magdalene, and St. Timothy in Ephesus.      All of these Early Christian pilgrimage centers produced primarily clay tubes for holy water; a total of only 43 lead tubes from this time period are known in the entire world, the Bulgarian archaeologists from the Burgas Museum point out.      They explaining 20 of those known lead tubes have been found in the St. John the Baptist Basilica in Monza, Italy (the Monza Cathedral); they were a gift from Lombard Queen Theodelinda (c. 570-628) made at the beginning of the 6th century.      Another 16 lead tubes have been found in a grave in the Bobbio Abbey (a monastery founded by Irish Saint Columbanus in 614 AD) in the Italian town of Bobbio, close to Milan.      One lead tube reliquary has been discovered in the Sant Pere de Casserres Abbey, a Benedictine monastery in the town of Les Masies de Roda, Osona comarca, Catalonia, Spain.      In addition to these lead tube reliquaries, three others are kept in Germany and four in the USA, all of which were produced in Jerusalem and have depictions of Gospel scenes.      Even though the reliquary discovered by the Bulgarian archaeologists in the basilica in the ancient and medieval fortress Burgos (Poros) on Cape Foros is also a lead tube, it is different from the other known lead tube reliquaries because the images on it are identical with the images from a group of clay tube reliquaries produced in ancient Ephesus.      Follow us on Facebook, Twitter, Google+, Tumblr!      “That is why at this stage we believe that the Burgas reliquary comes from this pilgrimage center (i.e. Ephesus) and it must be connected with the cult for St. John the Theologian (John the Apostle),” the head of the Burgas Museum of History, Milen Nikolov, explains.      He also notes that John the Apostle was particularly cherished by the Early Christians. According to the Bible, John was Jesus Christ’s favorite disciple, and when Jesus was crucified he asked John to take care of the Holy Mother, Virgin Mary.      Later, John the Apostle settled in the ancient city of Ephesus together with Virgin Mary and St. Mary Magdalene. This is where he wrote the Book of Revelation, also known as The Apocalypse, and lived till the rest of his life.      According to some historical sources, Christian pilgrims from around the world would gather on his grave in the Ephesus basilica on May 8, the date of his death. They would sprinkle rose petals on the rock above the basilica, and the next day wonder-working powder would appear on the rock. This manna could cure all kinds of diseases, which is why it was collected by the pilgrims in reliquaries and taken to their places of origin as evidence of their pilgrimage or as an apotropeus (an apotropaic item, i.e. an amulet chasing away evil).      Some scholars believe the manna collected by the pilgrims came from the pollen from the roses they placed on John the Apostle’s grave in Ephesus.      “That is why, at this point, we believe that a pilgrim from the fortress of Poros went on a pilgrimage to the grave of St. John the Theologian in Ephesus from where he brought the valuable reliquary with curing powder,” Nikolov elaborates.      The discovery of the lead tube reliquary with ashes from the grave of John the Apostle in Ephesus near Burgas resembles another relic discovery from the same region, Bulgaria’s Southern Black Sea coast.      Back in 2010 during excavations of an ancient monastery on the St. Ivan (St. John) Island in the Black Sea, off the coast of Bulgaria’s Sozopol, just to the north of Burgas (and the ancient and medieval port of Burgos (Poros) on Cape Foros), Bulgarian archaeologist Prof. Kazimir Popkonstantinov discovered a reliquary containing relics of St. John the Baptist. The relics of St. John the Baptist, which consist of small bone particles from a skull, jaw bone, arm bone, and tooth, have received lots of international interest in the years since then, and in February 2015 CNN reported that Oxford University scholars had confirmed the possibility of their authenticity by concluding that they belonged to a man who lived in the Middle East at the same time as Jesus Christ...
\\\bottomrule
    \end{tabular}
    \caption{Multi-News example. Document set with 2 individual input documents, separated by an \texttt{[END]} token. The comparison of summaries is in the following Table \ref{tab:qual_example2}. The input has been truncated to fit the page.}
    \label{tab:qual_example1}
\end{table*}

\begin{table*}[!ht]
\small
    \begin{tabular}{p{1.0\textwidth}}
        \toprule
\textbf{Reference} \\\midrule
Archaeologists in Bulgaria think they've found ashes belonging to one of the 12 apostles. The team found a small lead container with ashes while excavating a basilica that dates back to the sixth century AD, reports Archaeology in Bulgaria. The markings on the inch-long vessel match those at the grave of St. John the Apostle in what is now Turkey, leading researchers to surmise that a long-ago Christian from Bulgaria traveled to the grave site and returned home with the sample of ashes. Such journeys were common around that time, reports Novinite. “Probably a pilgrim from the Foros Peninsula went on a pilgrimage to Ephesus, and came back here with this relic, which was then donated to the basilica on Foros,” says the director of the Burgas Regional Museum of History in Bulgaria. (The ancient basilica is located in what is now the Black Sea city of Burgas.) Archeologists also found a Bulgarian royal seal from the 10th century at the dig site, which includes a fortress, notes the Week. (Another archaeological discovery: an ancient underground city in Turkey.) \\\midrule
\textbf{NLL} (47.05/16.12/24.59/28.10) \\\midrule
Archaeologists have made a pretty significant find in Bulgaria: ashes from the grave of one of Jesus Christ's Twelve Apostles. The lead tube containing the ashes of St. John the Theologian, who is believed to have lived in what is now Turkey, was found in a medieval fortress and is named for the early Christian basilica that once stood there, Ancient Origins reports. \textcolor{blue}{\textbf{The lead container, which is less than an inch long}}, is decorated with crosses and has only one handle, reports the International Business Times. The lead container is believed to have been carried by \textcolor{blue}{\textbf{a pilgrim who made the trip to the ancient city of Ephesus}}, where the Apostle was once a favorite of Jesus. The grave of John the Theologian was a popular pilgrimage destination for early Christians seeking healing, and the lead container "must be connected with the cult for St. John the Theologian," says the head of the Burgas Regional Museum of History. \textcolor{red}{\textbf{The find was announced Wednesday, but it wasn't made public until Thursday.}} (Click to read about how a lead container found in a 14th-century tomb could help heal ancient wounds.) \\\midrule
\textbf{ROUGE-L + RELAX} (56.76/25.06/30.76/37.10) \\\midrule
Archaeologists have made "one of the most important discoveries" in the history of the Burgas Regional Museum of History in Bulgaria, says the museum's director: a lead reliquary that contains ashes from the alleged grave of one of Jesus' Twelve Apostles. The \textcolor{blue}{\textbf{tiny lead container, which is less than an inch long}}, was found during last year's excavations of the fortress of Burgos on Cape Foros, reports Ancient Origins. The reliquary was once part of an early Christian basilica, and \textcolor{blue}{\textbf{"probably a pilgrim from the Foros Peninsula went on a pilgrimage to Ephesus, and came back here with this relic which was then donated to the basilica,"}} says Milen Nikolov. \textcolor{red}{\textbf{The find was announced Wednesday, but the news was made public on Wednesday.}} The find was made during the \textcolor{blue}{\textbf{same excavations that turned up a 10th-century royal seal}}, reports the Independent. The find is significant because early Christians believed the relic had healing properties, and John the Apostle's grave in Turkey was a popular pilgrimage site for those seeking healing, reports Ancient Origins. (Click to read about how a lead-tipped glass bottle was found in a 14th-century Roman city.) \\\midrule
\textbf{ROUGE-L + Coverage + RELAX} (57.59/30.00/31.93/38.64) \\\midrule 
Archaeologists have made "one of the most important discoveries" in the history of the Burgas Regional Museum of History in Bulgaria, says the museum's director: a lead reliquary that contains ashes from the alleged grave of one of Jesus Christ's apostles. The \textcolor{blue}{\textbf{tiny lead container, which is less than an inch long}}, was \textcolor{green(munsell)}{\textbf{found in a 6th-century basilica}} in what is now the \textcolor{blue}{\textbf{Black Sea city of Burgas}}, reports Ancient Origins. \textcolor{blue}{\textbf{"Probably a pilgrim from the Foros Peninsula went on a pilgrimage to Ephesus, and came back here with this relic which was then donated to the basilica}}," says Milen Nikolov, per the Independent. The reliquary is named for St. John the Theologian, who is believed to have been one of the Twelve Apostles. \textcolor{green(munsell)}{\textbf{The find was made during 2014 excavations at the fortress of Burgos on Cape Foros}}, but \textcolor{red}{\textbf{the announcement was made only this week}}. The head of the museum says the reliquary is connected to the \textcolor{green(munsell)}{\textbf{belief that there is manna, a holy curing powder, on the site of John the Apostle's grave every year}}. (Click to read about how a lead box found in a cave has been analyzed for ancient DNA.)
 \\
        \bottomrule
    \end{tabular}
    \caption{Multi-News example. Comparison of reference, NLL baseline, and RELAX-generated summaries for the document in Table \ref{tab:qual_example1}. We compare specifically the addition of the coverage term in the reward, to qualitatively show its importance. The R1/R2/RL/METEOR scores are shown in the headers. Highlighted in \textcolor{blue}{\textbf{blue}} are examples of key information that allow for the summary to remain faithful to the reference. Highlighted in \textcolor{green(munsell)}{\textbf{green}} are examples where the coverage term has managed to improve the quality of the summary. Highlighted in \textcolor{red}{\textbf{red}} are examples where the model has conveyed incorrect statements with respect to the input documents, and where the subsequent use of the coverage has seemingly improved it. We note that these results are also in line with the average scores presented in Table \ref{mn_results}.}
    \label{tab:qual_example2}
\end{table*}

\begin{table*}[!ht]
\small
    \begin{tabular}{p{1.0\textwidth}}
        \toprule
\textbf{Source Document} \\\midrule
'Greece's conservative prime minister-elect Kyriakos Mitsotakis vowed that the country would "proudly" enter a post-bailout period of "jobs, security and growth" after winning a landslide victory in Sunday's general election.  Official results showed Mitsotakis on track to crush leftist premier Alexis Tsipras, who oversaw austerity measures after Greece's dramatic rescue by international creditors in the European debt crisis.  "A painful cycle has closed," Mitsotakis said in a televised address, adding that Greece would "proudly raise its head again" on his watch.  "I will not fail to honour your hopes," he said as early congratulation calls came from outgoing European Commission chief Jean-Claude Juncker and Turkish President Recep Tayyip Erdogan.  With official results from 94 per cent of polling stations, New Democracy scored a crushing victory by nearly 40 per cent -- its best score in over a decade -- to 31.5 per cent for Tsipras's leftist Syriza party.  "I want to see this people prosper. I want to see the children who left to return," he later told party supporters.  Mitsotakis will be sworn in as Greece's new prime minister on Monday.  Tsipras had earlier admitted defeat after over four years in power that saw Greece emerge from its third bailout.  The 44-year-old warned that his Syriza party would "dynamically" resist efforts to scale back the party's pro-labour reforms.  If the results are confirmed, the 51-year-old Harvard graduate and former McKinsey consultant Mitsotakis will have a majority of 158 lawmakers in the 300-seat parliament. Tsipras's party will have 86 seats.  The final number will depend on how smaller parties fare. They need at least 3.0 percent of the vote to enter parliament.  New Democracy was last in power in 2014, in coalition with the Greek socialists.  Mitsotakis is a scion of one of Greece's top political families.  He is the son of former prime minister Constantine Mitsotakis, one of the country's longest-serving parliamentarians.  His sister is former minister Dora Bakoyannis, Athens's first female mayor. And new Athens mayor Costas Bakoyannis, elected in May, is his nephew.  Sunday's election was Greece's third in as many months, and the first held in midsummer since 1928.  In May, New Democracy beat Syriza by nearly 9.5 points in European parliament elections. A week later, it completed a near-sweep of Greek regions in local elections.  After that, Tsipras was forced to call an early general election. His term was scheduled to end in the autumn.  Greece's youngest premier in more than a century, Tsipras had trailed in the polls for months amid widespread dissatisfaction over high taxes.  "Greece is exiting 10 years of crisis and the new government will have the heavy task to give a chance to the country to recover completely or to sink", 36-year-old Aphrodite told AFP, as she cast her vote in the bohemian downtown Athens neighborhood of Exarcheia.  "I hope that from tomorrow we will be able to breathe with relief. To take a deep breath, if Mitsotakis does what he promises," added Athinodoros, a 48-year-old self-employed worker. Tsipras has accused Mitsotakis -- who was part of a 2012-2014 crisis government -- of "disastrous" mismanagement that brought hundreds of thousands of job losses and business failures.  Mitsotakis has now pledged to create "better" jobs through growth, foreign investment and tax cuts and to "steamroll" obstacles to business.  Tsipras -- who reduced unemployment and raised the minimum wage for the first time since 2012 -- was criticized for campaigning as an anti-austerity crusader before eventually accepting a third EU bailout and the economic cutbacks that entailed.  In parts of the country, there was also a backlash against a controversial agreement with North Macedonia that ended a bitter 27-year dispute over the country's name.  The new smaller parties fighting to secure representation are Greek Solution, a nationalist party formed by TV salesman Kyriakos Velopoulos, and MeRA25, an anti-austerity party founded by maverick economist and former Greek finance minister Yanis Varoufakis.  According to the exit polls, Varoufakis's party could elect nine lawmakers.  Greek Solution could end up with 10 deputies, while neo-Nazi party Golden Dawn looks likely to be shut out of parliament for the first time since 2012.  Golden Dawn, until recently Greece's third-ranking party, is in steep decline amid an ongoing trial for the 2013 murder of an anti-fascist rapper, allegedly carried out with the knowledge of senior Golden Dawn members. Mitsotakis has promised to hit the ground running. A Eurogroup finance meeting on Monday will convene to discuss the state of Greece's economy after tax cuts rolled out by Tsipras in May.  Get Breaking news, live coverage, and Latest News from India and around the world on NDTV.com. Catch all the Live TV action on NDTV 24x7 and NDTV India. Like us on Facebook or follow us on Twitter and Instagram for latest news and live news updates. Budget 2019: Find the latest news on ndtv.com/budget. Use the income tax calculator to learn about your tax liability \texttt{[END]} Investors expect new Greek Prime Minister Kyriakos Mitsotakis to prove that his business-friendly reputation is deserved.  The former banker and management consultant will need to make good on pledges to address issues including government finances, soured loans and crippling bureaucracy, while working within tight fiscal constraints. Although he has inherited an economy on the mend and a stock market that is soaring, they are rebounding from shrunken bases. Mr Mitsotakis must ensure that Greece can attract the investment it desperately needs and create jobs as the country digs itself out of a financial crisis that has lasted more than a decade and taken a toll on living standards. Here are the three main issues the new Greek government will have to deal with from day one:  While the new government is not yet in place, the country's creditors want to send a clear message that it has to stick to its commitment of achieving a 3.5 per cent primary surplus every year until 2022. Former prime minister Alexis Tsipras' move to distribute handouts before the European elections has raised doubts about Greece's ability to meet its fiscal targets.  The European Commission estimates that the freebies will lead to a fiscal cost of 1 per cent of gross domestic product for both this year and the next, meaning creditors may ask the new government for additional austerity measures. Mr Mitsotakis plans to rapidly legislate tax cuts that will come into effect from next year to spur economic activity and show investors that Greece is creating a more friendly business environment.  The biggest challenge is addressing about €80 billion (S\$122 billion) in bad loans. Lenders are speeding up efforts to cut soured debt by selling portfolios of non-performing exposures (NPEs), but they will need more tools to meet their ambitious targets of single-digit NPE ratios by 2021.  Mr Mitsotakis' target is doubling Greece's growth rate to 4 per cent next year. To achieve that, he needs investments. To convince investors that they can trust the country again, he wants to immediately proceed with the long-delayed Hellinikon project.  The flagship venture envisages the transformation of the former Athens airport site - more than two times the size of New York's Central Park - into a metropolitan park including luxury hotels, casino, marinas and apartments.  But that will not be enough. The new government will have to deal with red tape, a sluggish judicial system and corruption, as well as speeding up privatisations, especially in the energy sector. \texttt{[END]}...
\\\bottomrule
    \end{tabular}
    \caption{WCEP example. Document set with 25 individual input documents, separated by an \texttt{[END]} token. The comparison of summaries is in the following in Table \ref{tab:qual_example4}. The input has been truncated to fit the page.}
    \label{tab:qual_example3}
\end{table*}

\begin{table*}[!ht]
\small
    \begin{tabular}{p{1.0\textwidth}}
        \toprule
\textbf{Reference} \\\midrule
Winner of the general election Kyriakos Mitsotakis is sworn in as the new Prime Minister of Greece, succeeding Alexis Tsipras. \\\midrule
\textbf{NLL} (20.00/7.14/20.00/5.26) \\\midrule
Greek voters go to the polls for a general election. \\\midrule
\textbf{ROUGE-L + RELAX} (70.58/68.75/70.58/56.68) \\\midrule
Conservative politician \textcolor{blue}{\textbf{Kyriakos Mitsotakis is sworn in as the new Prime Minister of Greece}}. \\\midrule
\textbf{ROUGE-L + Coverage + RELAX} (68.18/57.14/63.63/58.02) \\\midrule 
Conservative politician \textcolor{blue}{\textbf{Kyriakos Mitsotakis is sworn in as the new Prime Minister of Greece}} after \textcolor{green(munsell)}{\textbf{defeating leftist leader}} \textcolor{blue}{\textbf{Alexis Tsipras}} in \textcolor{green(munsell)}{\textbf{yesterday's election}}. \\
        \bottomrule
    \end{tabular}
    \caption{WCEP example. Comparison of reference, NLL baseline, and RELAX-generated summaries for the document in Table \ref{tab:qual_example3}. We compare specifically the addition of the coverage term in the reward, to qualitatively show its importance. The R1/R2/RL/METEOR scores are shown in the headers. Highlighted in \textcolor{blue}{\textbf{blue}} are examples of key information that allow for the summary to remain faithful to the reference. Highlighted in \textcolor{green(munsell)}{\textbf{green}} are examples where the coverage term has managed to improve the quality of the summary. As mentioned in Section \ref{sec:results}, shorter summaries are involved in this dataset, and are more likely to result in higher ROUGE scores. In this example, both RELAX objectives have drastically improved the accuracy. We can also see that the model has been able to use the coverage term to improve the summary quality by adding relevant fragments, and lead to a higher METEOR score. We note that these results are in line with the average scores presented in Table \ref{wcep_results}.}
    \label{tab:qual_example4}
\end{table*}

\end{document}